\documentclass{article}

\usepackage{iclr2027_conference,times}
\usepackage{amsmath,amsfonts,bm}
\usepackage{hyperref}
\usepackage{url}
\usepackage{graphicx}
\usepackage{wrapfig}
\usepackage{placeins}
\usepackage{tabularray}
\usepackage{booktabs}
\usepackage{multirow}
\usepackage{colortbl}
\iclrfinalcopy

\usepackage{amsmath,amsfonts,bm}

\def\eqref#1{equation~\ref{#1}}

\def\1{\bm{1}}

\DeclareMathAlphabet{\mathsfit}{\encodingdefault}{\sfdefault}{m}{sl}
\SetMathAlphabet{\mathsfit}{bold}{\encodingdefault}{\sfdefault}{bx}{n}

\DeclareMathOperator*{\argmin}{arg\,min}

\title{FaceLinkGen: A Re-evaluation of Identity Leakage in Privacy-Preserving Face Recognition and Face Anonymization Systems Using Simple Distillation}

\author{Wenqi Guo, Qingyun Qian, Mohamed S. Shehata, and Shan Du \\
Department of Computer Science \\
University of British Columbia \\
\texttt{\{wg25r, qingyunq\}@student.ubc.ca} \\ 
\texttt{\{mohamed.sami.shehata, shan.du\}@ubc.ca}}

\begin{document}

\maketitle

\begin{abstract}
Privacy-preserving face recognition (PPFR) and face anonymization have different goals, but both must retain some identity-related information for their intended use. We show that an adaptive attacker can learn this information. We propose FaceLinkGen, a simple distillation-based attack that trains a face recognition model to map protected inputs back to standard face embeddings. FaceLinkGen applies to keyless PPFR systems and perception-preserving face de-identification (De-ID) systems. For PPFR, the recovered embeddings can be used to regenerate faces that match the original person. Across MinusFace, PartialFace, and DecoyFace, the regenerated faces achieve acceptance rates of 81.0--99.4\% on Face++ and 74.9--99.6\% on Amazon. For De-ID, FaceLinkGen links protected faces to unprotected images of the same person, reaching Recall@1 values of 48.4--89.6\% in the one-side-protected setting across the evaluated methods. FaceLinkGen exposes identity leakage across all evaluated methods, including DecoyFace and WDP, whose protection resists the tested U-Net attacks on face recovery and protected-to-unprotected linkage, respectively. The attack also remains effective when trained with limited paired data. Code is available at \url{https://anonymous.4open.science/r/FaceLinkGenRelease-08C8/}. 
\end{abstract}

\raggedbottom
\section{Introduction}

Two common facial privacy tasks are privacy-preserving face recognition (PPFR) and perception-preserving face anonymization or de-identification (De-ID). PPFR aims to let a server verify a user's identity without exposing the user's face \citep{erkin_privacypreserving_2009}, while perception-preserving De-ID has a nearly opposite goal: the protected face should remain recognizable to humans while becoming unrecognizable to face recognition (FR) models.

Representative PPFR methods include frequency-domain approaches such as PartialFace \citep{mi_privacypreserving_2023}, MinusFace \citep{mi_privacypreserving_2024}, and FracFace~\citep{dai_fracface_2025}, as well as decoy-oriented approaches such as DecoyFace \citep{ren_decoyface_2026}. The De-ID methods studied in this paper include the perturbation-based TIP-IM \citep{yang_face_2021} and Protego \citep{wang_protego_2025}, synthesis-based PerceptFace \citep{wang_make_2025}, and the diffusion-based \textit{Weakening Diffusion Purification} (WDP) \citep{salar_enhancing_2025}.

Both tasks face a common tension: privacy-coupled identity information must remain available to support their intended utility. In a keyless PPFR system, no user-held secret key controls access to the protected representation. If that representation remains useful for identity matching, an attacker with query access can learn to extract its identity signal and regenerate a visually recognizable face. Similarly, if a perception-preserving De-ID image must still look like the same person to humans, it necessarily retains identity cues that an FR model can learn to use.

\begin{figure}
    \centering
    \includegraphics[width=0.9\linewidth]{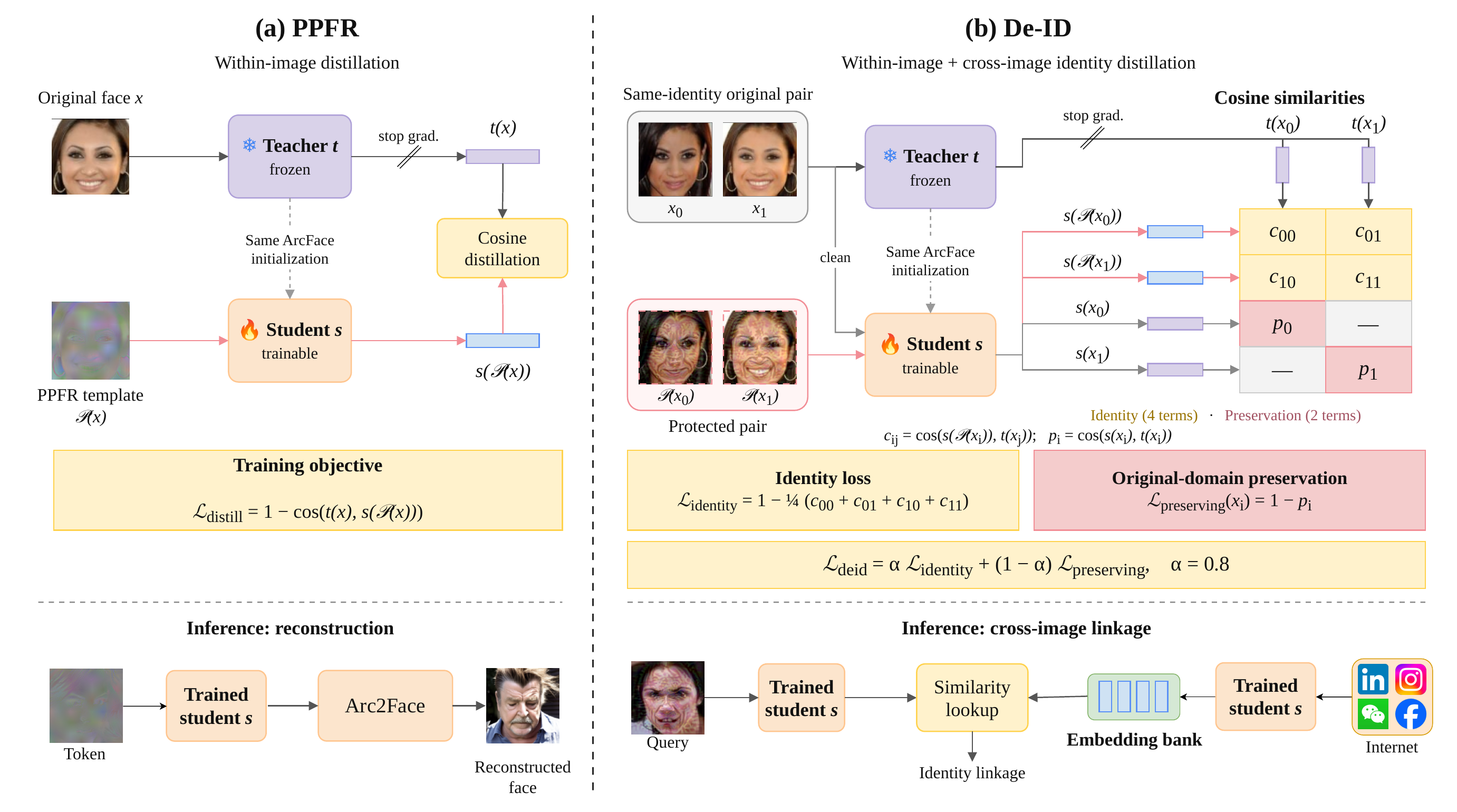}
    \caption{Method overview of FaceLinkGen. (a) For PPFR, training extracts the ground-truth identity embedding from the original face and encourages the student to produce the same embedding from the protected template by matching the teacher's output using a cosine-similarity loss. At inference time, the student extracts an embedding from an unseen template, and Arc2Face is used to synthesize a face image from that embedding. (b) For De-ID systems, we use two images of the same identity along with their protected counterparts. The goal is to maximize cosine similarity across the $2 \times 2$ matching grid between the two original images and their protected counterparts. To preserve the student's ability to recognize unprotected faces, we also feed the student an original image and enforce that its output for that image matches the teacher's.}
    \label{fig:method}
\end{figure}
We propose FaceLinkGen, a unified distillation-based attack framework for PPFR and De-ID systems. Its main advantage is coverage across all evaluated protection methods. Our results show that the vulnerability is not merely theoretical. For PPFR, FaceLinkGen regenerates faces with acceptance rates of 81.0--99.4\% on Face++ and 74.9--99.6\% on Amazon across MinusFace, PartialFace, and DecoyFace. For De-ID, FaceLinkGen achieves 48.4--89.6\% Recall@1 when matching a query against a gallery (with either the query or the gallery protected) across the evaluated methods, demonstrating that identity linkage remains practical after protection. Although an improved attack (U-Net) achieves a slightly higher attack success rate (ASR) on MinusFace, PartialFace, and TIP-IM, FaceLinkGen can recover identities from DecoyFace and link WDP-protected queries to unprotected images, where the tested U-Net attacks fail.

In summary, our contributions are as follows: (1) we identify a shared structural weakness underlying keyless PPFR and perception-preserving face de-identification, two tasks with seemingly opposite goals; (2) we propose FaceLinkGen, a unified distillation-based attack framework that exploits the identity signal retained for utility; and (3) we show that substantial identity information survives the protection mechanisms of representative methods from both families, with successful attacks across all evaluated methods, including methods that resist the tested U-Net attacks.

\section{Related Work}
\paragraph{Privacy-Preserving Face Recognition}
The aim of PPFR is to enable a server to confirm a user's identity without revealing the user's face. For example, a person could use facial authentication to approve a transaction on an online banking site, while ensuring that the bank does not receive the original image and cannot reconstruct the user's face from the provided template. The unprotected face never leaves the user's device, and the conversion happens locally. The server will only receive the protected template. 

Most existing PPFR approaches work in the frequency domain. Their main assumption is that machines recognize faces using high-frequency information, whereas the visual information perceived by humans lies in the low-frequency domain. On this basis, they argue that the original image cannot be reconstructed after the low-frequency information is removed and the frequency channels are disrupted. Among them, PartialFace first removes the low-frequency information and then shuffles the high-frequency information, whereas MinusFace uses an autoencoder-style network to reconstruct the original image and computes the residual between the original and reconstructed images to remove low-frequency information. The residual thus contains very little information. MinusFace then shuffles it again in the frequency domain and converts it back to the spatial domain. Another method, FracFace, uses a fractal-style transformation to disrupt the frequency channels. In contrast, DecoyFace separates reconstruction-sensitive information from recognition-sensitive information and replaces the reconstruction-sensitive information with another person's face. A trained reconstruction network therefore recovers the decoy face rather than the protected identity.

\paragraph{Face Anonymization}
There are two types of face anonymization or de-identification (De-ID): utility-preserving and non-utility-preserving. The simplest non-utility-preserving method is blurring the face. A more advanced non-utility-preserving method is a form of ``face swap,'' where a different face replaces the user's face to protect privacy. This is useful for dataset sharing and anonymizing people other than the main subject (e.g., protecting passersby's privacy). Utility-preserving De-ID methods aim to keep the face recognizable to humans while avoiding recognition by FR models. For example, users may want to share photos on social media that friends and family recognize, while preventing web scrapers from indexing their faces to build a face search engine that links their photos across platforms. In this paper, we target utility-preserving methods, referred to simply as ``face De-ID'' in the following text.

There are two main types of such De-ID systems: adversarial perturbation-based methods and perception-preserving synthesis-based methods. Adversarial perturbation-based methods add noise to exploit model weaknesses and avoid recognition. However, these perturbations are often weak and model-specific \citep{wang_make_2025}, and they introduce visible artifacts. Perception-preserving synthesis-based methods like PerceptFace generate a different identity while keeping visual changes minimal, so the face still appears to belong to the same person to human observers.

\paragraph{Previous Attacks}
Prior work by \citet{calmon_fundamental_2015} shows that perfect privacy is possible only if the information needed for utility is decoupled from private information. This principle underlies our attack: human perception and machine recognition rely on largely shared facial information, making full separation difficult, if not impossible.

Recent work in non-facial domains has also begun shifting attention from pixel-level reconstruction to semantic-level inversion. In such settings, the attacker aims to regenerate information that is semantically consistent with the original identity without recovering the original image itself. This goal is easier to achieve and better aligned with realistic attack objectives. For instance, \citet{yue_gradient_2023} propose a semantic recovery framework that uses generative models to synthesize semantically consistent images without pixel-level similarity in federated learning attacks. Other findings suggest that traditional reconstruction metrics fail to capture how humans perceive privacy leakage~\citep{sun_privacy_2023}.

\section{Threat Model}
\paragraph{Threat Model for PPFR} For PPFR systems, the adversary is the service provider itself. The service provider does not need to be the algorithm developer, which can be a third party or a standardization agency. This follows the original threat model of PPFR \citep{erkin_privacypreserving_2009}, where the user wants the server to verify identity without seeing the real face. We model the server as honest-but-curious (HBC) following prior work \citep{erkin_privacypreserving_2009,ren_decoyface_2026}: it faithfully follows the protocol (without compromising any systems) but attempts to recover the user's facial image from the templates it receives. We restrict the adversary to black-box oracle access to the protection function: it can submit an original image and obtain the corresponding protected template, but cannot inspect internal states or random seeds; any internal randomness in the method is freshly sampled for each query and is unknown to the adversary. 

This constraint includes two scenarios: first, when the service provider did not build the protection mechanism and lacks knowledge of the system; and second, when an insider, such as an employee, can access the data and invoke the protection function but cannot view the system design. It also extends to an external attacker who can run the PPFR method locally on their own device to obtain the pairs, although this attacker is not our primary adversary. 

This matches the black-box setting of MinusFace \citep{mi_privacypreserving_2024} and is stricter than the ``minimal assumptions'' setting of \citet{zhang_validating_2024}, which requires thousands of server queries per target and models the adversary more like an outside attacker than a service provider. 

We consider two regimes. In the first, the HBC service provider can query the protection method without rate limits, which applies when the protection method is an open standard or developed by the service provider itself. In the second, the service provider has limited access to the protection method, so the adversary obtains templates for far fewer identities (256--8192). This tests attack performance when paired data are limited, such as when the protection method is held by a third-party service or is closed-source.

\paragraph{Threat Model for De-ID} For De-ID systems, the adversary is a web scraper that collects online images, stores their embeddings in a database, and searches for matching faces. The adversary's goal is to link a query face to other images of the same person, whether those images are protected or not. Because several De-ID methods require per-sample optimization, making large-scale data generation costly for an attacker, we evaluate training with 2{,}048 images from 1{,}024 identities, as well as a few-sample setting with 32--1{,}024 identities. We also assume a black-box setting in which the attacker can acquire these pairs by directly querying the De-ID tools using public face datasets such as MS1M without knowledge of system internals, since these tools must be made publicly available for users to anonymize their faces.

We justify our threat models by Kerckhoffs's principle, which states that security should not rely on keeping an algorithm secret; rather, it should rely on keeping the key secret. Thus, the adversaries do not have access to the random components in the systems but do have access to an implementation of the algorithm itself. 



\section{FaceLinkGen} 

\paragraph{Background}


For a face image $x$ and its protected image $\tilde{x}$, we assume that they contain model-identifiable information $z_m$ and human-perceptual information $z_p$. PPFR methods aim to remove $z_p$ while retaining $z_m$, so that the image remains usable for recognition but reveals nothing to a human observer and cannot be reconstructed back to $x$. De-ID methods aim for the opposite: to hide or alter $z_m$ while retaining $z_p$, so that the image is still identifiable to a human but cannot be matched by a recognition system. Both directions assume that $z_m$ and $z_p$ can be separated; however, in natural face images, we hypothesize that they are strongly correlated, such that retaining one leaves enough identity signal for an adaptive model to learn a mapping back to the other. We tested this hypothesis using our proposed FaceLinkGen attack framework. 

Consider PPFR first. Methods such as Arc2Face~\citep{papantoniou_arc2face_2024} and FaceID IP-Adapter~\citep{ye_ipadapter_2023a} can regenerate a face from $z_m$ (the face embedding) alone. The fact that this works demonstrates the core issue: $z_m$ carries enough identity information to render a face recognizable to humans. Given that PPFR retains $z_m$, it is reasonable to construct an attack that extracts Arc2Face-compatible $z_m$ and generates a new image of the person, without directly extracting $z_p$.


De-ID follows two routes, and both fail for related reasons. The first adds adversarial perturbations that mask $z_m$, so that an off-the-shelf model $\mathcal{F}$ produces $\mathcal{F}(\tilde{x}) \neq z_m$. Here $z_m$ is still present in the image, and only the model's reading of it has been disrupted. Training with the objective 
\(
\argmin_{\mathcal{F}} \mathcal{L}(\mathcal{F}(\tilde{x}), z_m)
\)
recovers the mapping. This is similar to adversarial training, where models trained on adversarial examples can defend against adversarial attacks, at least for the specific methods they were trained on \citep{goodfellow_explaining_2015}.

The second route, exemplified by PerceptFace, applies a transformation that replaces $z_m$ in $x$ with another person's identity, so $z_m$ is genuinely changed rather than hidden. This appears stronger, but it carries a structural constraint: for the method to be useful, the result must still look like the same person to a human observer, which means $z_p$ must be preserved. If a human can still recognize the person using $z_p$, then $z_p$ retains enough information for re-identification. We can therefore train a model to learn a mapping from $z_p$ back to $z_m$ embeddings and use them for linkage.


\paragraph{Our Method}

Figure~\ref{fig:method} summarizes FaceLinkGen. We initialize a student model $s$ and a teacher model $t$ from the same pre-trained ArcFace feature extractor with identical weights. To maintain a unified framework and ensure a fair comparison, we use the Arc2Face-compatible model even for De-ID tasks, which do not require regeneration. The teacher $t$ remains frozen throughout training, while only the parameters of $s$ are updated.

Given a public face dataset $\mathcal{X}$ with an example image $x \in \mathcal{X}$ and a protection transformation $\mathcal{P}$, the protected input is $\tilde{x} = \mathcal{P}(x)$. The central objective is to train $s$ to recover from $\tilde{x}$ the identity representation produced by $t$ on the corresponding original image. We use cosine distance as the distillation objective:
$$
\mathcal{L}_{\mathrm{distill}} = 1 - \cos\bigl(s(\mathcal{P}(x)),\, \operatorname{sg}(t(x))\bigr),
$$
where $\operatorname{sg}$ denotes the stop-gradient operator.

For PPFR, single-image distillation is sufficient because the attack objective is to recover an identity representation from an individual protected template. The student therefore learns the mapping $$\mathcal{P}(x) \xrightarrow{s} t(x)$$.

For De-ID systems, however, single-image supervision alone is insufficient. The teacher embedding $t(x)$ may encode image-specific factors such as pose, illumination, and expression, which the protected image retains, unlike in PPFR. If the student is trained only to reproduce the teacher embedding of the exact corresponding image, it may exploit instance-specific correlations rather than learning an identity-consistent representation. Additionally, since the attacker's goal here is to link the image with \emph{another} image of the same person in the database, cross-image supervision is also better aligned with the inference setting.

We therefore introduce cross-image supervision. Given two images $(x_0, x_1)$ of the same identity, each protected image is matched against the teacher embeddings of both original images:
$$
\mathcal{L}_{\mathrm{identity}} = 1 - \frac{1}{4}\sum_{i=0}^{1}\sum_{j=0}^{1}\cos\bigl(s(\mathcal{P}(x_i)),\, \operatorname{sg}(t(x_j))\bigr).
$$

The diagonal terms $(i=j)$ provide within-image distillation, while the off-diagonal terms $(i\neq j)$ explicitly encourage identity consistency across different images of the same person.

A realistic De-ID adversary may encounter a mixture of protected and unprotected images. The student should therefore recover identity from protected inputs without degrading its representation of ordinary faces. To preserve performance on the original image domain, we introduce an original-domain-preserving loss:
$$
\mathcal{L}_{\mathrm{preserving}} = 1 - \cos\bigl(s(x),\, \operatorname{sg}(t(x))\bigr).
$$

The final PPFR training objective is $\mathcal{L}_{\mathrm{PPFR}} = \mathcal{L}_{\mathrm{distill}}$. For De-ID, we combine identity recovery and original-domain preservation. Since the preserving loss only acts as a regularization term to maintain the model's existing capabilities, we assign it a lower weight:

$$
\mathcal{L}_{\mathrm{DeID}} = \alpha \mathcal{L}_{\mathrm{identity}} + (1-\alpha)\mathcal{L}_{\mathrm{preserving}}, \qquad \alpha = 0.8.
$$

At inference, FaceLinkGen supports different downstream attacks depending on the protection setting. For PPFR, suppose an HBC service provider receives an unknown victim's protected template $T$, where $T \notin \{ \mathcal{P}(x_i) : x_i \in \mathcal{X}\}$. The trained student first maps $T$ into the ArcFace identity space as $e_T = s(T)$. The recovered embedding can then condition Arc2Face to synthesize a face corresponding to the victim:

$$
F = \operatorname{Arc2Face}(e_T, \epsilon), \qquad \epsilon \sim p(\epsilon),
$$

where $\epsilon$ is the diffusion initialization noise that controls identity-independent attributes not uniquely specified by the face embedding, such as pose and expression.

For De-ID, the attacker instead uses the recovered representation for identity linkage. Let $\mathcal{C}$ denote a large-scale corpus of public face images, which may contain either protected or unprotected images. The attacker first constructs an embedding gallery: $\mathcal{G} = \{(s(x),\, x) : x \in \mathcal{C}\}.$

Given a protected query face $F$, its recovered embedding is $e_F = s(F)$. The attacker then identifies the gallery image whose embedding has the highest cosine similarity to the query:
\[
x^{*} = \arg\max_{x \in \mathcal{C}}\, \cos\bigl(e_F,\, s(x)\bigr).
\]

This links the protected face to an identity represented in the public image corpus.

Because the recovered embedding is task-independent, the De-ID linkage attack also applies to PPFR, and the PPFR reconstruction attack also applies to De-ID. These cross-applications provide limited practical advantage. PPFR is not intended to conceal identity from recognition systems, since identity information must be preserved for authorization. Conversely, reconstructing a visually recognizable face from a De-ID representation is of limited value because the protected face already preserves the human-perceptual identity information of the original face. These cross-applications could nevertheless be useful when a PPFR system is intended only for within-system verification and should not enable cross-dataset linkage, or when an image regenerated from a protected De-ID image could serve as a token for linkage in a database with a different embedding space. These cases are outside the scope of this paper.

\paragraph{Metrics}
For PPFR systems, where the goal is to recover human-recognizable faces, we use Arc2Face \citep{papantoniou_arc2face_2024} to reconstruct images from distilled embeddings. To evaluate reconstruction fidelity, we verify generated faces against originals using both Face++ (\texttt{FAR@1e-5}) and Amazon (binary pass/fail output) verification APIs, two APIs popular in the face privacy research community, under their strictest operational thresholds. We also report the AdaFace cosine similarity between original and reconstructed faces, and present uncurated qualitative examples. Note that Face++, the Amazon API, and AdaFace are \emph{not} our attack targets; they are used as proxies for human visual judgment to verify that the generated and original faces depict the same person. 

\begin{wrapfigure}{r}{0.5\textwidth}
    \centering
    \includegraphics[height=0.2\textwidth]{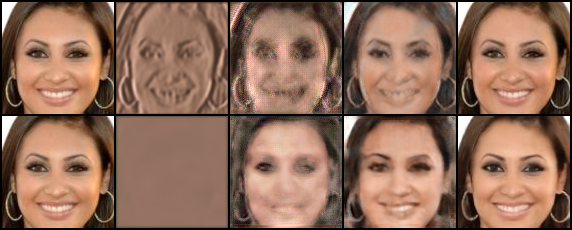}
    \caption{U-Net attack results for PartialFace (top row) and MinusFace (bottom row). IN denotes instance normalization. From left to right, the columns show the original image, SGD training without IN, SGD training with IN at an earlier training step, SGD training with IN at a later training step, and AdamW+IN training.}
    \label{fig:unet_minus_frac}
    \vspace{-0.5cm}
\end{wrapfigure}

For De-ID systems, since the goal is to prevent face-based retrieval, we evaluate retrieval performance on a closed-set face gallery under two settings: both gallery and query images are protected (the setting that is harder for the defender per \citet{wang_protego_2025}), and only one randomly selected side (either query or gallery) is protected (the setting that is harder for the attacker per \citet{wang_protego_2025}). We report Recall@1, Recall@5, and mean reciprocal rank (MRR).

\paragraph{Targets and Baseline} 
We evaluate three PPFR methods: the frequency-based MinusFace \citep{mi_privacypreserving_2024} and PartialFace \citep{mi_privacypreserving_2023}, and the decoy-based DecoyFace \citep{ren_decoyface_2026}. For MinusFace and PartialFace, we use official implementations from \texttt{Tencent/TFace}. Because public DecoyFace code is unavailable, we reproduced it using Paper-Replication \citep{hans_codingagents_2026} and verified the key metrics against the original results (Table~\ref{tab:reprod} in the Appendix). Our reproduced values are close to those reported in the original paper. We omit FracFace \citep{dai_fracface_2025} because there are inconsistencies between the code and the paper, and neither their code nor our reproduced code can replicate its results. For De-ID systems, we evaluate the perturbation-based TIP-IM \citep{yang_face_2021} and Protego \citep{wang_protego_2025}, the perception-based PerceptFace \citep{wang_make_2025}, and the diffusion-based Weakening Diffusion Purification (WDP) \citep{salar_enhancing_2025}. We focus on dodging methods rather than impersonation methods such as CLIP2Protect \citep{shamshad_clip2protect_2023}. Impersonation methods can make an FR model identify a ``protected'' image as someone else, but they do not prevent the model from also recognizing the person's actual identity, offering little privacy protection.

Given that most facial datasets are collected from public images and often contain images of the same people across datasets, and it is difficult to deduplicate identities across datasets, using different datasets for training and testing actually has a higher risk of data leakage than using a single dataset and splitting the images by person ID. Thus, we use only MS1M \citep{guo_msceleb1m_2016} as the real image dataset in our paper. To further validate how well our method transfers to a different distribution, we use TPDNE (\texttt{TLeonidas/this-person-does-not-exist}) as a synthetic dataset for PPFR testing in Table~\ref{tab:tpdne} in the appendix. Since TPDNE does not have paired samples, we cannot use it for De-ID evaluation.

For PPFR, we use 30{,}000 identities for training, 2{,}000 for validation, and 2{,}000 for testing. For De-ID systems, where per-sample optimization makes sample generation computationally expensive (both in our experiments and for real attackers), we use 2{,}048 training images across 1{,}024 identities, 128 validation identities, and 384 test identities. Both settings split the dataset based on identity labels and not image indices. Closed-set retrieval evaluation uses 192 query images and 192 target images, mixed with 107{,}484 unprotected distractor images, reflecting realistic web-scale retrieval where only a small minority of photos are protected.

As a baseline, we evaluate U-Net \citep{ronneberger_unet_2015} inversion, which trains a network to directly reconstruct clean faces from protected inputs using pixel-level loss. This architecture represents the standard reconstruction attack used in prior evaluations of MinusFace, PartialFace, FracFace, and DecoyFace. We apply it to both PPFR and De-ID systems. For PPFR, the U-Net is trained with an $\ell_1$ reconstruction loss: \(\mathcal{L}_{\text{ppfr-unet}} = \lVert x - \mathcal{U}(\mathcal{P}(x)) \rVert_1.\) For MinusFace and PartialFace, we evaluate both the original architecture from \citet{mi_privacypreserving_2024} (without normalization) and an improved version incorporating instance normalization \citep{ulyanov_instance_2017}. The former reproduces the results reported in the original paper, in which the authors claimed that the protection method was successful, and the latter gives results under a better-tuned attacker. For DecoyFace, following \citet{ren_decoyface_2026}, we use an asymmetric U-Net that maps intermediate feature representations to RGB images. 

For De-ID systems, since the attacker also faces a mixture of original and protected images, the U-Net also includes an identity-preserving loss similar to that used in our method, with $\beta = 0.8$: \(\mathcal{L}_{\text{deid-unet}} = \beta \cdot \lVert x - \mathcal{U}(\mathcal{P}(x)) \rVert_1 + (1 - \beta) \cdot \lVert x - \mathcal{U}(x) \rVert_1.\)

\begin{figure}[t] 
    \centering
    \includegraphics[height=0.23\textwidth]{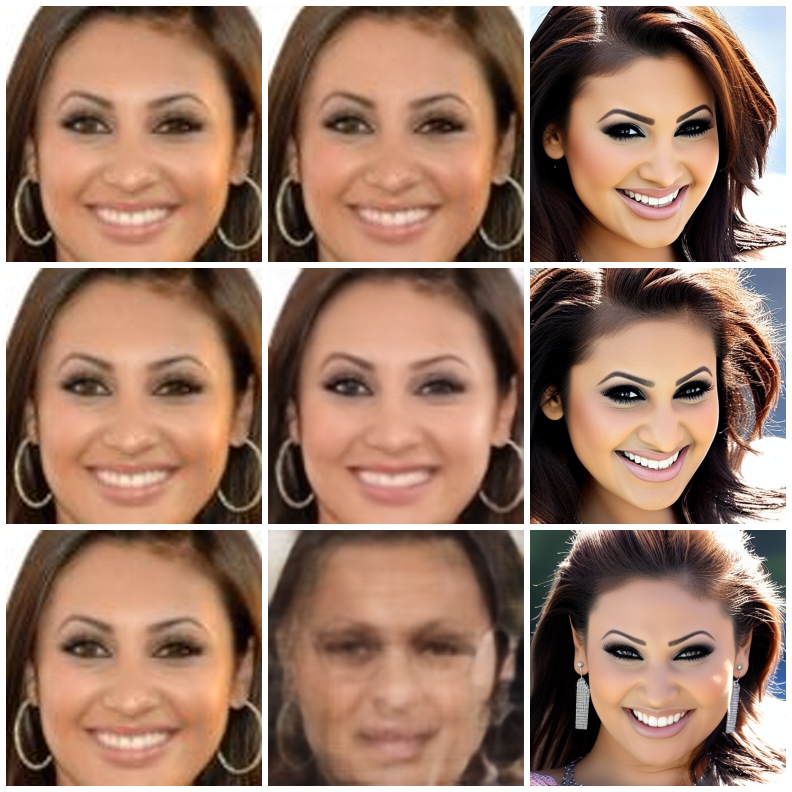} \hspace{10pt}
    \includegraphics[height=0.23\textwidth]{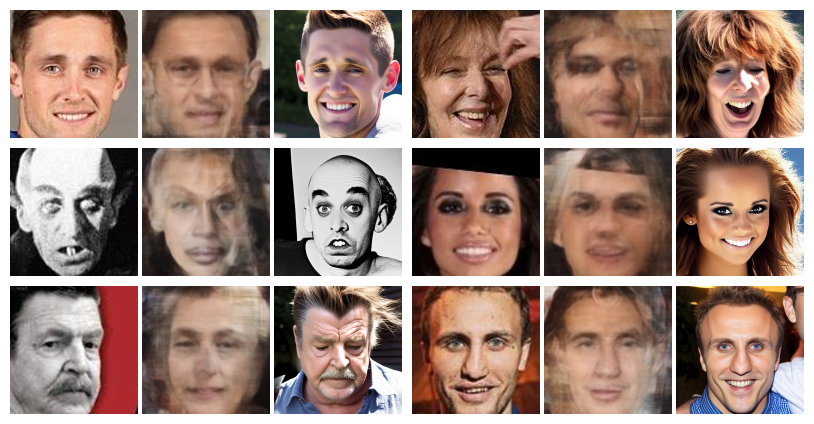}
    \caption{Left: Reconstructions produced by each attack. Rows correspond to PartialFace, MinusFace, and DecoyFace; columns show the original image, the U-Net attack, and the distillation attack. Right: More uncurated examples of reconstruction comparisons between U-Net and distillation for DecoyFace.}
    \label{fig:decoy}
    \vspace{-0.2in}
\end{figure}


\section{Results}
\subsection{PPFR Systems}

\paragraph{Revisiting the U-Net Attack} We first re-examine U-Net inversion against PartialFace and MinusFace. Although MinusFace and PartialFace reported that these representations resist U-Net inversion, our better-trained U-Net successfully recovers clean faces (Figure~\ref{fig:unet_minus_frac}, rightmost column). Our experiments attribute this discrepancy to architectural and optimization choices. Training the original MinusFace U-Net (without normalization layers) using SGD \citep{sutskever_importance_2013} produces nearly uniform images where inversion fails. This differs from the blurry faces reported in the original papers, but in both cases the inversion clearly fails. Adding instance normalization improves feature propagation: early in training, coarse facial structures emerge while fine details remain blurry, matching reported behavior in the prior papers; with continued training, structures become clearer, but the details are still hard to recognize.

\begin{table}
  \footnotesize
  \centering
  \caption{PPFR method results under U-Net and distillation attacks. Both AdamW+IN U-Net and distillation successfully invert MinusFace and PartialFace, achieving high acceptance rates on Face++ and Amazon and exceeding the AdaFace similarity threshold (0.19, selected to maximize F1 on LFW). Distillation is the only evaluated attack that succeeds against DecoyFace.}
  \label{tab:ppfr_methods}
  \begin{tabular}{llccc}
  \toprule
  Attack Method                   & Metric      & MinusFace & PartialFace & DecoyFace \\
  \midrule
  \multirow{3}{*}{AdamW+IN U-Net} & Face++      & 0.998     & 1.000       & 0.014     \\
                                  & Amazon      & 0.998     & 1.000       & 0.000     \\
                                  & AdaFace Sim & 0.803     & 0.940       & 0.074     \\
  \midrule
  \multirow{3}{*}{SGD U-Net}      & Face++      & 0.000     & 0.008       & --     \\
                                  & Amazon      & 0.000     & 0.064       & --    \\
                                  & AdaFace Sim & 0.009     & 0.268       & --    \\
  \midrule
  \multirow{3}{*}{Distillation (Ours)}
                                  & Face++      & 0.980& 0.994& 0.810     \\
                                  & Amazon      & 0.994& 0.996& 0.749     \\
                                  & AdaFace Sim & 0.629& 0.696& 0.457     \\
  \bottomrule
  \end{tabular}
  \vspace{-0.5cm}
\end{table}

Switching from SGD to AdamW alongside instance normalization enables complete reconstruction. This indicates that the apparent resilience of prior defenses reflected evaluation against under-converged U-Nets rather than inherent resistance to inversion. On DecoyFace, however, U-Net reconstructions depict the decoy identity rather than the original subject, confirming that decoy substitution successfully misleads direct pixel inversion. This matches the results reported in the DecoyFace paper.

\paragraph{FaceLinkGen Results} 

Figure~\ref{fig:decoy} (left) compares reconstructions from the tuned U-Net (AdamW+IN) and FaceLinkGen. Table~\ref{tab:ppfr_methods} reports the attack success rates (ASR) verified by commercial Face++ and Amazon APIs, as well as the AdaFace cosine similarity between original and reconstructed faces. Under the original MinusFace training setup (SGD without normalization), U-Net inversion fails across both MinusFace and PartialFace. With AdamW and instance normalization, however, U-Net inversion achieves near-100\% verification rates on MinusFace and PartialFace. FaceLinkGen achieves Face++ acceptance rates of 98.0\% and 99.4\% on them as well, comparable to 99.8\% and 100.0\% for AdamW+IN U-Net. Its AdaFace similarity is slightly lower because FaceLinkGen performs semantic reconstruction, generating another image of the same person, unlike U-Net inversion, which recovers the original image. However, an AdaFace similarity of 0.6--0.7 is well within same person differnt image range. FaceLinkGen's main advantage is that it also breaks DecoyFace (81.0\% on Face++, 74.9\% on Amazon), which completely resists U-Net inversion. 

On DecoyFace, even AdamW-trained U-Nets cannot pass face verification: although the reconstructions resemble faces, they depict the decoy subject, which is the intent of DecoyFace. In contrast, FaceLinkGen achieves an 81.0\% verification rate on Face++ and 74.9\% on Amazon, recovering most of the identities. Uncurated visual examples in Figure~\ref{fig:decoy} (right) further support this: although U-Net can reconstruct a face, it does not match the original person's identity, whereas FaceLinkGen can generate another face of the same individual. This demonstrates that while decoy substitution diverts U-Net reconstruction, the template retains sufficient identity information for recognition, which distillation reliably extracts and uses to regenerate faces.

To verify embedding quality directly, we compare cosine similarity between student and teacher embeddings against the empirical intra-identity similarity distribution (Figure~\ref{fig:dist} in Appendix). The student-teacher similarity exceeds the similarity between different photos of the same individual, confirming that distilled embeddings faithfully preserve the identity signal more than another image of the same person would. 

\subsection{De-ID Systems}
Table~\ref{tab:deid} reports the De-ID attack results. We use the one-side-protected setting as our primary measure of attack effectiveness because in the both-sides-protected setting, retrieving one protected image using another protected image of the same individual is already highly effective before any attack. For example, PerceptFace achieves 0.917 Recall@1 before attack. This high baseline recall stems from two factors. First, because the gallery consists predominantly of unprotected photos (as on the real Internet, where only a small minority of users protect their photos), protection shifts embeddings into distinct regions rather than hiding them in the unprotected crowd, making protected images stand out and cluster together; t-SNE visualizations (Figure~\ref{fig:tsne} in Appendix~\ref{app:embeddings}) confirm that all evaluated methods except Protego form isolated embedding clusters, which corresponds to their high both-sides-protected recall. Second, several methods deterministically transform facial features rather than removing them, effectively assigning each person a consistent pseudonym across photos. This mirrors classic linkage flaws, such as using a virtual but stable MAC address to avoid device tracking, static pseudo-identifiers, or unsalted hashes: while the original value is masked, deterministic substitution leaves records linkable across sessions, even though the true identity is never recovered.

\begin{table}[t]
\centering
\scriptsize
\setlength{\tabcolsep}{3pt}
\caption{Attack results on the De-ID systems: recall on the 192-ID test pool (gallery size 107{,}676). ``Average'' is the arithmetic mean of the scores from the one-side and both-sides settings, reflecting a realistic scenario where protected and unprotected images coexist in the gallery.}
\label{tab:deid}
\begin{tabular}{ll ccc ccc ccc}
\toprule
 & & \multicolumn{3}{c}{U-Net} & \multicolumn{3}{c}{Distillation (Ours)} & \multicolumn{3}{c}{Before Attack} \\
\cmidrule(lr){3-5} \cmidrule(lr){6-8} \cmidrule(lr){9-11}
 & & Recall@1 & Recall@5 & MRR & Recall@1 & Recall@5 & MRR & Recall@1 & Recall@5 & MRR \\
\midrule
\multirow{3}{*}{TIP-IM}
 & One side  & 0.969 & 0.974 & 0.973 & 0.896 & 0.958 & 0.923 & 0.031 & 0.052 & 0.045 \\
 & Both sides & 0.974 & 0.979 & 0.975 & 0.885 & 0.922 & 0.905 & 0.255 & 0.391 & 0.317 \\
 & \cellcolor{gray!15}Average & \cellcolor{gray!15}\textbf{0.971} & \cellcolor{gray!15}\textbf{0.977} & \cellcolor{gray!15}\textbf{0.974} & \cellcolor{gray!15}0.891 & \cellcolor{gray!15}0.940 & \cellcolor{gray!15}0.914 & \cellcolor{gray!15}0.143 & \cellcolor{gray!15}0.222 & \cellcolor{gray!15}0.181 \\
\midrule
\multirow{3}{*}{WDP}
 & One side  & 0.031 & 0.042 & 0.036 & 0.484 & 0.703 & 0.578 & 0.005 & 0.021 & 0.011 \\
 & Both sides & 0.490 & 0.667 & 0.577 & 0.620 & 0.745 & 0.673 & 0.682 & 0.839 & 0.762 \\
 & \cellcolor{gray!15}Average & \cellcolor{gray!15}0.260 & \cellcolor{gray!15}0.354 & \cellcolor{gray!15}0.306 & \cellcolor{gray!15}\textbf{0.552} & \cellcolor{gray!15}\textbf{0.724} & \cellcolor{gray!15}\textbf{0.625} & \cellcolor{gray!15}0.344 & \cellcolor{gray!15}0.430 & \cellcolor{gray!15}0.387 \\
\midrule
\multirow{3}{*}{PerceptFace}
 & One side  & 0.255 & 0.438 & 0.342 & 0.823 & 0.911 & 0.858 & 0.036 & 0.062 & 0.052 \\
 & Both sides & 0.891 & 0.932 & 0.911 & 0.911 & 0.943 & 0.925 & 0.917 & 0.974 & 0.938 \\
 & \cellcolor{gray!15}Average & \cellcolor{gray!15}0.573 & \cellcolor{gray!15}0.685 & \cellcolor{gray!15}0.627 & \cellcolor{gray!15}\textbf{0.867} & \cellcolor{gray!15}\textbf{0.927} & \cellcolor{gray!15}\textbf{0.891} & \cellcolor{gray!15}0.477 & \cellcolor{gray!15}0.518 & \cellcolor{gray!15}0.495 \\
\midrule
\multirow{3}{*}{Protego}
 & One side  & 0.542 & 0.677 & 0.607 & 0.880 & 0.932 & 0.908 & 0.026 & 0.026 & 0.027 \\
 & Both sides & 0.349 & 0.474 & 0.415 & 0.911 & 0.943 & 0.926 & 0.026 & 0.047 & 0.039 \\
 & \cellcolor{gray!15}Average & \cellcolor{gray!15}0.445 & \cellcolor{gray!15}0.576 & \cellcolor{gray!15}0.511 & \cellcolor{gray!15}\textbf{0.896} & \cellcolor{gray!15}\textbf{0.938} & \cellcolor{gray!15}\textbf{0.917} & \cellcolor{gray!15}0.026 & \cellcolor{gray!15}0.036 & \cellcolor{gray!15}0.033 \\
\bottomrule
\end{tabular}
\vspace{-0.5cm}
\end{table}

For the attack results shown in Table~\ref{tab:deid}, both the U-Net attack and our distillation attack achieve meaningful ASR against TIP-IM and PerceptFace by retrieving the target person from the gallery. FaceLinkGen extends attack coverage to WDP in the one-side-protected setting, where U-Net reaches only 0.031 Recall@1 and distillation reaches 0.484. Even though the number 0.484 might seem low, it means that around half of the tested users' identities are compromised, which indicates substantial leakage. Distillation also achieves substantially higher recall than U-Net on PerceptFace (0.823 vs. 0.255) and Protego (0.880 vs. 0.542), although U-Net achieves higher Recall@1 than distillation on TIP-IM in both settings (0.969/0.974 vs. 0.896/0.885), likely because the weak protection of TIP-IM causes ASR for both methods to saturate around 0.9--1.0. We also show in the Appendix that even with only a small set of samples, our method can attack both types of systems.


\section{Conclusion}

We introduced FaceLinkGen, a unified distillation attack targeting both keyless PPFR and perception-preserving face De-ID systems. FaceLinkGen reconstructs human-recognizable faces from PPFR templates and can link De-ID queries to unprotected images. In many scenarios, FaceLinkGen matches U-Net's ASR and surpasses it in others. It also remains effective in cases where U-Net fails entirely (DecoyFace). Overall, these results indicate that existing keyless PPFR approaches and De-ID systems are vulnerable to distillation attacks. We therefore call on future developers to carefully consider the robustness of privacy-preserving face systems against distillation attacks before deployment. We have discussed more potential defense methods in the Appendix. 

\section{AI Usage Disclosure}
LLMs were used to search for related work, brainstorm ideas, proofread, and implement ideas by writing code and monitoring experiments. A human was always in the loop to monitor the LLMs, exercise critical judgment, and make the final decisions about their outputs.

\bibliography{main,missing}
\bibliographystyle{iclr2027_conference}
\appendix
\clearpage
\section{Embedding Visualization}
\label{app:embeddings}

\begin{figure}[htbp]
    \centering
    \includegraphics[width=0.6\textwidth]{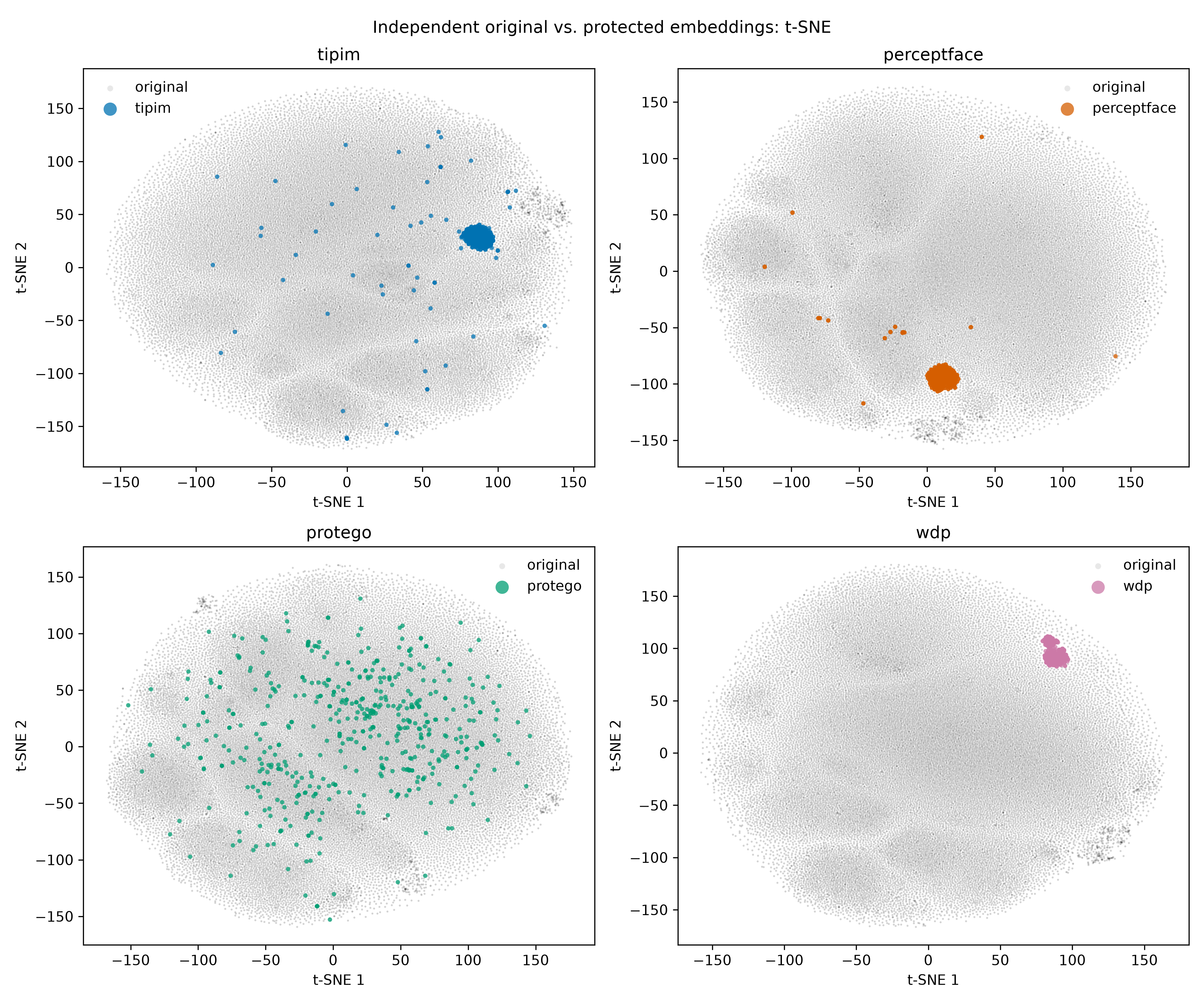}
    \caption{t-SNE plot of each method's embeddings compared to the original embeddings. Gray dots represent embeddings of unprotected distractor images.}
    \label{fig:tsne}
\end{figure}


\FloatBarrier
\section{Implementation Details}

The protection methods produce different output formats. All De-ID systems produce RGB images, whereas the evaluated PPFR methods produce arrays of different shapes. MinusFace and PartialFace both produce $C \times H \times W$ tensors with different values of $C$, while $H$ and $W$ remain the original image height and width. We replace the first convolutional layer of the student model with one that accepts $C$ input channels rather than three, without changing the remaining architecture. DecoyFace produces intermediate CNN feature cubes (by default, $64 \times 56 \times 56$). To adapt to this output, we remove layers from the student model until the input shape of the next layer is similar to that of DecoyFace, then adjust the channel count with a single convolutional adapter and the spatial dimensions with bilinear interpolation. Because the student and DecoyFace models use the same Res-100 \citep{he_deep_2015} architecture, the relevant feature tensors have the same shape, and the convolutional adapter is only for compatibility if the model changes. Although they share an architecture, the models come from different families (AdaFace \citep{kim_adaface_2023} vs. ArcFace \citep{deng_arcface_2022}) to avoid circularity and an artificial boost in attack performance. In the full-data setting, all distillation methods are trained using a learning rate of \texttt{1e-4} and weight decay of 0.01 with AdamW \citep{loshchilov_decoupled_2019}. All runs are executed on an NVIDIA DGX B200. Training on PPFR methods is stopped when it plateaus, while De-ID training is stopped when the validation loss rises again (indicating overfitting), and the checkpoints with the lowest validation loss are selected due to the smaller sample size and greater training instability.

\begin{table}[!htbp]
    \centering
    \caption{Reproduction results compared with those reported in the original DecoyFace paper. Our reproduction performs better on three of the four metrics, with a lower identity leakage ratio, a higher identity redirection ratio, and a higher face validity ratio. It has slightly higher cosine similarity with the original face (0.081), but the value remains within the near-zero range of similarities between different people, and thus the Identity Leakage Ratio remains low.}
    \label{tab:reprod}
    \begin{tabular}{ccc}
        \toprule
        Metric & Reproduced Result & Original Paper Result \\
        \midrule
        Cosine similarity ($\downarrow$) & 0.081 & \textbf{0.027} \\
        Identity Leakage Ratio ($\downarrow$) & \textbf{2.61}\% & 2.93\% \\
        Identity Redirection Ratio ($\uparrow$)& \textbf{97.39}\% & 97.07\% \\
        Face Validity Ratio ($\uparrow$) & \textbf{99.99}\% & 99.88\% \\
        \bottomrule
    \end{tabular}
\end{table}

\FloatBarrier
\section{Soft Biometric Extraction}
Beyond identity recognition, exposure of soft biometric attributes presents a significant privacy risk. Characteristics such as skin color, age, and gender are sensitive personal data that enable unauthorized profiling and algorithmic discrimination. This threat model applies only to PPFR, which aims to identify a person while preserving their privacy. De-ID systems deliberately preserve human-recognizable appearance, so retaining soft biometrics is intentional. Privacy frameworks like the Canadian Privacy Act \citep{branch_consolidated_2025} explicitly protect race and age. A robust privacy-preserving system must therefore prevent the recovery of these attributes from its templates.

The regenerated faces show that these soft biometrics remain visible. This agrees with prior findings that ArcFace embeddings retain such information \citep{melzi_multi-ive_2023, osorio-roig_reversing_2023}. Because our extracted embeddings closely resemble the original facial embeddings, we hypothesize that a model can learn a direct mapping from the template to these attributes without facial reconstruction.

To assess this, we appended a lightweight probe to the trained student model and trained the resulting model for 10 epochs on the FairFace dataset \citep{karkkainen_fairface_2019} to predict age, gender, and race (7 categories). The probe consists of a shared hidden layer with ReLU activation, followed by three task-specific linear output heads. The model was trained with cross-entropy loss, a learning rate of \texttt{2e-5}, and weight decay of 0.01 using AdamW. To establish a baseline for soft biometric extraction, we also trained the same probe architecture on raw images, using the undistilled teacher model as the base model. The results are reported in Table~\ref{tab:soft}.

\begin{figure}[!htbp]
    \centering
    \includegraphics[width=0.6\linewidth]{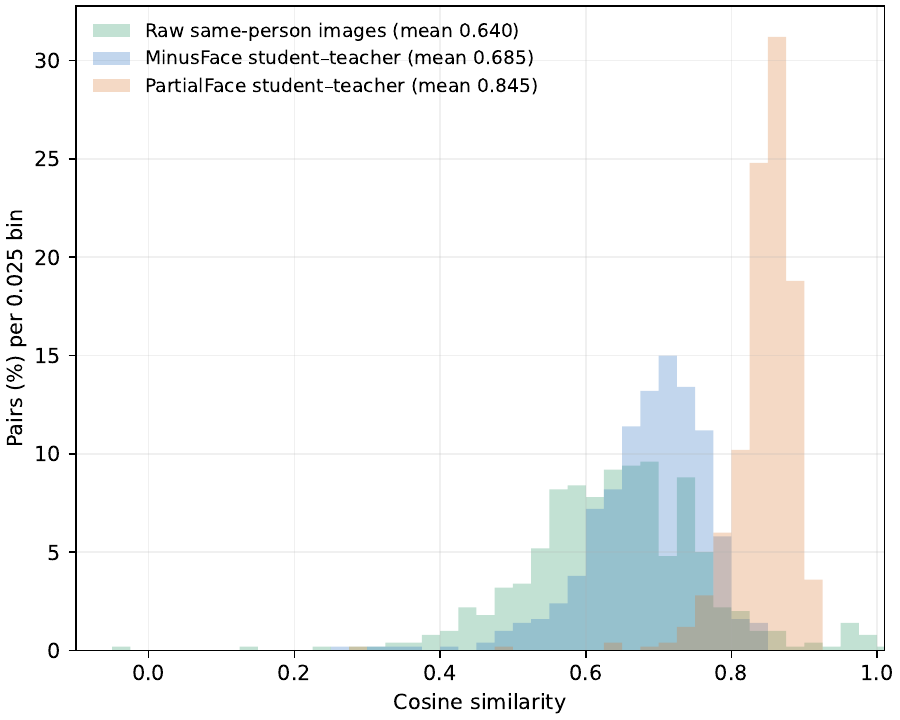}
    \caption{Distributions of cosine similarity for raw same-person image pairs and student-teacher embedding pairs from MinusFace and PartialFace.}
    \label{fig:dist}
\end{figure}

\begin{table}
  \centering
  \caption{PPFR method results under U-Net and distillation attacks, evaluated against a different image from the same identity directory. MinusFace and PartialFace distillation use the best checkpoints after continued training. The teacher baseline uses embeddings of the original unprotected inputs. The teacher baseline is shown in separate rows because it is not associated with any protection method. }
  \label{tab:ppfr_methods_diffimg}
  \begin{tabular}{llccc}
  \toprule
  Attack Method & Metric & MinusFace & PartialFace & DecoyFace \\
  \midrule
  \multirow{2}{*}{AdamW+IN U-Net} & Amazon & 0.962 & 0.986 & 0.000 \\
   & AdaFace Sim & 0.519 & 0.600 & 0.041 \\
  \midrule
  \multirow{2}{*}{SGD U-Net} & Amazon & 0.000 & 0.030 & -- \\
   & AdaFace Sim & 0.007 & 0.163 & -- \\
  \midrule
  \multirow{2}{*}{Distillation (Ours)} & Amazon & 0.784 & 0.890 & 0.419 \\
   & AdaFace Sim & 0.425 & 0.470 & 0.307 \\
  \midrule
  \multirow{2}{*}{Teacher baseline} & Amazon & \multicolumn{3}{c}{0.926} \\
   & AdaFace Sim & \multicolumn{3}{c}{0.493} \\
  \bottomrule
  \end{tabular}
  \par\smallskip
  \begin{minipage}{\linewidth}
  \end{minipage}
\end{table}
\begin{table}[!htbp]
\centering
\small
\caption{Soft biometric recovery performance from the protected template without regeneration. Acc denotes accuracy, and MAE denotes mean absolute error in years. Age MAE is calculated using the midpoint of each age group, with ages greater than 70 marked as 75.}
\label{tab:soft}
\begin{tabular}{lllll}
\toprule
            & Gender Acc & Race Acc & Age Group Acc & Age MAE  \\
\midrule
MinusFace   &92.52\%& 64.97\%& 54.43\%&5.28\\
PartialFace &93.49\%&68.64\%&56.55\%&4.92\\
DecoyFace   & 89.39\%    & 62.20\%  & 53.43\%       & 5.66     \\
Raw Images  & 94.29\%    & 71.28\%  & 60.30\%       & 4.40      \\
\bottomrule
\end{tabular}
\end{table}

\section{Few-Sample Settings}
\label{sec:few}
\begin{figure}
    \centering
    \scriptsize
    \includegraphics[width=\linewidth]{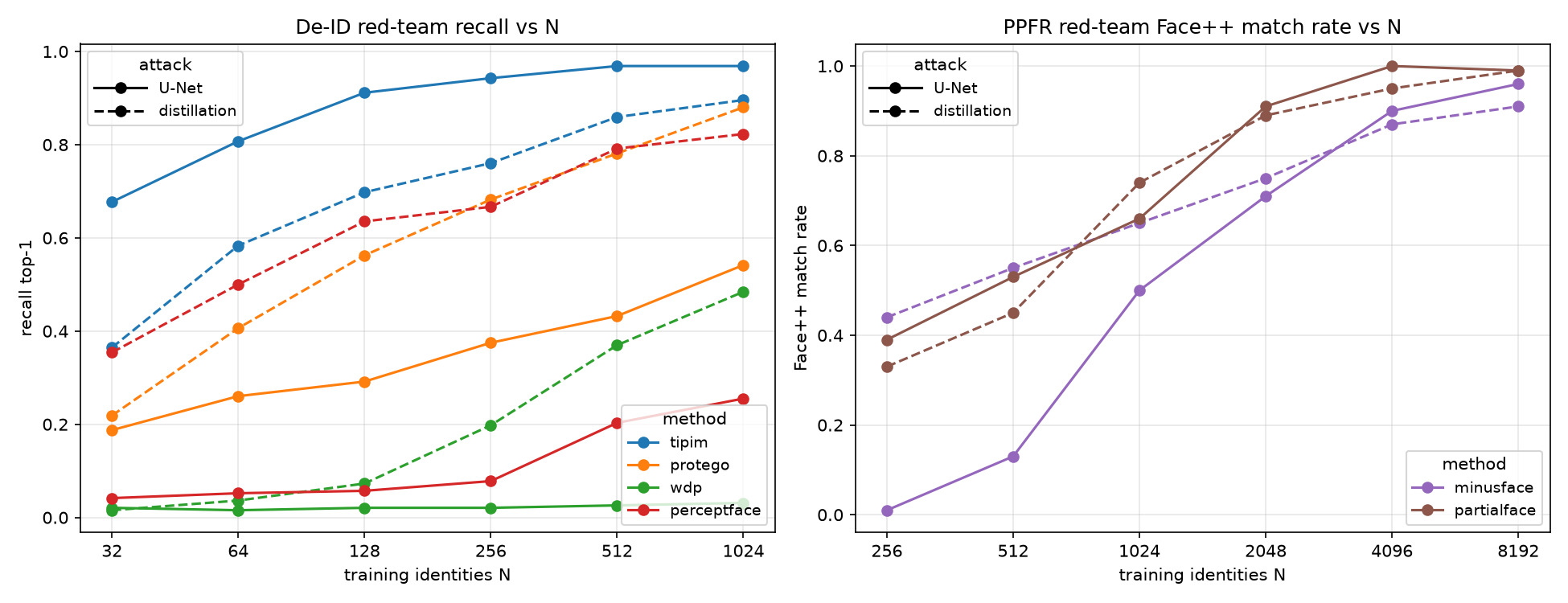}
    \caption{Few-sample attack success rate as a function of $N$. The De-ID ASR is the average Recall@1 across the one-side-protected and both-sides-protected settings, and the PPFR attack success rate is defined as the rate of successful verification by Face++. $N$ refers to the number of identities.}
    \label{fig:few_sample}
\end{figure}
We also evaluate attacks in a few-sample regime, where the adversary can acquire only a limited number of training pairs. Prior studies such as PerceptFace \citep{wang_make_2025} observe that inversion is possible with large paired datasets and propose rate limiting as a countermeasure. We therefore investigate whether attacks succeed under strict data budgets. We train both distillation and U-Net models with a learning rate of \texttt{5e-5}, since both are fine-tuned, across $N$ identities with 2 images per identity, evaluating $N \in \{256, 512, 1024, 2048, 4096, 8192\}$ for PPFR and $N \in \{32, 64, 128, 256, 512, 1024\}$ for De-ID. These identity counts reflect realistic constraints: PPFR service providers typically have broader but still limited query access, whereas De-ID adversaries are usually outsiders operating under tighter query budgets.

Because the distillation student is pre-trained for face recognition, comparing it against a U-Net trained from scratch would disadvantage the U-Net in data-limited regimes. To ensure a balanced comparison, we pre-train the U-Net on clean faces using an autoencoder reconstruction loss: \[\mathcal{L} = \lVert x - \mathcal{U}(x)\rVert_1.\] We temporarily omit skip connections during initial pre-training to prevent trivial identity shortcuts. The U-Net is trained on the full MS1M dataset, as this pre-training does not use protected data pairs, using AdamW with a learning rate of \texttt{1e-3} until validation loss plateaus, after which skip connections are re-enabled and training continues until convergence. This pre-trained model initializes all few-sample U-Net experiments.

Figure~\ref{fig:few_sample} plots the attack success rate against the number of training identities. For most De-ID methods, the distillation attack achieves substantially higher ASR than the U-Net does. On WDP, U-Net yields an ASR of nearly zero, whereas distillation reaches 0.40. TIP-IM is the only exception, for which U-Net outperforms distillation, as TIP-IM's protection is weak: its simple additive perturbations are readily removed by standard denoising. Notably, distillation on TIP-IM follows roughly the same curve as on other methods. For PPFR methods, both U-Net and distillation can successfully attack PartialFace with relatively high ASR even with few training identities. For MinusFace, however, distillation maintains a much higher ASR than the U-Net does until both are nearly saturated at around 2{,}048 identities.


\begin{table}
  \centering
  \caption{PPFR attacks on 200 aligned synthetic faces from the TPDNE dataset. Each reconstructed image is compared with the original cropped input from which the protected representation was generated. Amazon and Face++ report acceptance rates; AdaFace reports mean cosine similarity.}
  \label{tab:tpdne}
  \begin{tabular}{llrrr}
  \toprule
  Attack & PPFR method & Amazon & Face++& AdaFace Sim \\
  \midrule
  AdamW+IN U-Net & MinusFace & 1.000 & 1.000 & 0.722 \\
  AdamW+IN U-Net & PartialFace & 1.000 & 1.000 & 0.908 \\
  AdamW+IN U-Net & DecoyFace & 0.005 & 0.015& 0.060 \\
  \midrule
  Distillation & MinusFace & 0.985 & 0.980& 0.515 \\
  Distillation & PartialFace & 1.000 & 1.000 & 0.638 \\
  Distillation & DecoyFace & 0.415 & 0.660& 0.333 \\
  \bottomrule
  \end{tabular}
\end{table}

\section{Future Defense Directions}
We suggest several potential directions for future PPFR design and evaluation, focusing mainly on stronger defense mechanisms and broader vulnerability assessments.

One rigorous approach is to incorporate secret keys into the conversion process, as in \citet{yuan_generating_2022}. This works as a multi-factor authentication system that requires both biometrics and a key, and it prevents attackers, including our method, from converting a face without the secret. A method could transmit the face embedding instead of the face itself, since there are many ways to incorporate keys into embeddings, such as rotation and dropout \cite{zhong_slerpface_2024}.

Alternatively, systems may return to formal cryptographic methods such as \cite{ao_cryptoface_2025}. These methods are traditionally viewed as computationally expensive, but modern computing resources make this trade-off acceptable. For instance, \citet{jindal_secure_2020} reports a processing time of only 2.83\,ms per face pair. Importantly, these computational costs can act as an effective client-side constraint against brute-force attacks, similar in concept to slow hashing. This enhances privacy while remaining imperceptible to regular users.

For De-ID tasks, although our evidence is not strong enough to show that retaining information for human perception while preventing model identity extraction is impossible, our results suggest that this is a likely possibility, and future work should weigh this trade-off carefully. Possible workarounds include using a similar key-based structure in which a person’s face is altered to a different identity when posting on social media by a reversible transformation, and only trusted viewers can restore it to the original user's face when viewing.
\end{document}